\documentclass[runningheads]{llncs}

\usepackage[american]{babel}
\usepackage[utf8x]{inputenc}

\usepackage[long]{optional}
\usepackage{todonotes} % use option [disable] in final version

\usepackage{graphicx}
\usepackage{mathptmx}
\usepackage{booktabs}
\usepackage{tabularx}
\usepackage[export]{adjustbox}
\usepackage[ruled,vlined]{algorithm2e}
\usepackage{amsmath,amssymb,amsfonts}
\usepackage[hidelinks,pdftex,unicode]{hyperref}
\usepackage{url}
\usepackage{csquotes}
\usepackage[shortlabels,inline]{enumitem}

\usepackage[capitalize,noabbrev]{cleveref}
\Crefname{equation}{Formula}{Formulae}

\begin{document}

\title{Negation in Cognitive Reasoning\thanks{The authors gratefully
acknowledge the support of the German Research Foundation (DFG) under the grants
SCHO~1789/1-1 and STO~421/8-1 \emph{CoRg -- Cognitive Reasoning}.}}


\author{Claudia Schon\inst{1}\orcidID{0000-0003-2455-0974} \and
	Sophie Siebert\inst{2}\orcidID{0000-0002-6812-796X} \and
	Frieder Stolzenburg\inst{2}\orcidID{0000−0002−4037−2445}}

\institute{
Universität Koblenz-Landau,
Institute for Web Science and Technologies,
Universitätsstr.~1,
56070~Koblenz, Germany,
\email{schon@uni-koblenz.de},
\url{http://www.uni-koblenz.de/}
	\and
Harz University of Applied Sciences,
Automation and Computer Sciences Department,
Friedrichstr.~57--59,
38855 Wernigerode, Germany,
\email{\{ssiebert,fstolzenburg\}@hs-harz.de},
\url{http://artint.hs-harz.de/}}

\maketitle

\begin{abstract}
Negation is both an operation in formal logic and in natural language by which a
proposition is replaced by one stating the opposite, as by the addition of
\enquote{not} or another negation cue. Treating negation in an adequate way is
required for cognitive reasoning, which aims at modeling the human ability to
draw meaningful conclusions despite incomplete and inconsistent knowledge. One
task of cognitive reasoning is answering questions given by sentences in natural
language. There are tools based on discourse representation theory to convert
sentences automatically into a formal logic representation, and additional
knowledge can be added using the predicate names in the formula and knowledge
databases. However, the knowledge in logic databases in practice always is
incomplete. Hence, forward reasoning of automated reasoning systems alone does
not suffice to derive answers to questions because, instead of complete proofs,
often only partial positive knowledge can be derived, while negative knowledge
is used only during the reasoning process. In consequence, we aim at eliminating
syntactic negation, strictly speaking, the negated event or property. In this
paper, we describe an effective procedure to determine the negated event or
property in order to replace it by its inverse. This lays the basis of cognitive
reasoning, employing both logic and machine learning for general question
answering. We evaluate our procedure by several benchmarks and demonstrate its
practical usefulness in our cognitive reasoning system.

\end{abstract}

\section{Introduction}

Negation is a very common operation in natural language. It reverses the
meaning of a statement or parts of it. Especially in text comprehension, it is
important to correctly identify and process negation because negation can strongly
alter the meaning of the overall contents of a statement. In this paper, we
focus on solving commonsense reasoning tasks in natural language, which requires
an adequate handling of negation.

Commonsense reasoning is the sort of everyday reasoning humans typically perform
about the world \cite{Mul14}. It needs a vast amount of implicit background
knowledge, such as facts about objects, events, space, time, and mental states.
While humans solve commonsense reasoning tasks relatively easily and
intuitively, they are rather difficult for a computer.

Most approaches that tackle commonsense reasoning and its benchmarks do not use any
reasoning at all and rely mainly on machine learning and statistics on natural
language texts. These often very successful approaches combine
unsupervised pre-training on very large text corpora with supervised learning procedures
\cite{DBLP:conf/naacl/DevlinCLT19,RN+18}. However, the good results
currently achieved, e.g., with BERT \cite{DBLP:conf/naacl/DevlinCLT19} should be
critically questioned. \cite{niven-kao-2019-probing} presents a system based on
BERT that is only slightly below the average untrained human baseline on the
Argument Reasoning Comprehension Task \cite{habernal-etal-2018-semeval}, but a close
examination of the learned model by the authors of \cite{niven-kao-2019-probing} shows that there are statistical
cues in the dataset that have nothing to do with the represented problems.
The good performance of BERT is explained by BERT exploiting these
cues. Examples of those cues are the words \enquote{not}, \enquote{is},
\enquote{do}, \enquote{are}, \enquote{will not} and \enquote{cannot}, with
\enquote{not} being the strongest cue. After creating a version of the dataset
without the bias of the cue \enquote{not}, the authors find that performance of BERT drops
to the random baseline. This illustrates that the system based on BERT in
\cite{niven-kao-2019-probing} has no underlying
understanding of the tasks. This observation emphasizes the importance of
explanations in commonsense reasoning and negation handling.

Therefore, we take a different way and rely on background knowledge
represented in logic together with automated reasoning (cf. \cite{SSS19b}). In a
first step, the natural language representation of a benchmark problem is
translated into a first-order logic formula using the KnEWS system \cite{knews}
-- a tool based on discourse representation theory \cite{KR93}.
Since many of
the benchmark problems contain negations, it is important that they are treated
correctly. Taking a look back at the
detected bias from \cite{niven-kao-2019-probing}, it becomes clear that negation
in particular is responsible for the majority of the biases. The word
\enquote{not} is not only the strongest cue, but other cues are compound of it.
This requires increased caution when dealing with negations.

By eliminating the word \enquote{not}, additional technical advantages apply:
While KnEWS can successfully detect negation, it often generates formulae with a
large negation scope. Often it spans a whole sentence or
subphrase. Our goal is to localize the negation as precisely as possible and
thus to pinpoint the exact negated event or property, which can later be
replaced by its non-negative counterpart or antonym -- the inverse of the
negated event or property. In many cases, the negation can even be completely
removed from the formula in this way, which often enables an automated reasoning system
to derive more positive knowledge. In the CoRg system \cite{SSS19a} (cf. \cref{sec:corg}),
we use this knowledge as input for a neural network calculating a
likelihood for each answer alternative to fit to the respective premise.
A precisely determined (scope of the) negation in the formal representation also
erases ambiguity and facilitates further processing.

In this paper, we therefore present an approach to automatically identify a negated event
or property using a natural language sentence and the corresponding logical formula
generated by KnEWS. \cref{sec:back} describes negation in logic as well as
natural language and its importance in cognitive reasoning, which aims at
modeling the human ability to draw meaningful conclusions despite incomplete and
inconsistent knowledge \cite{FH+19a}. In \cref{sec:meth}, we describe the
cognitive reasoning system CoRg, in which this work is embedded, including
challenges caused by negation. In \cref{sec:eval}, we present experimental
results and demonstrate the usefulness of our approach for cognitive reasoning.
\cref{sec:final} contains a summary, conclusions, and an outlook for future work.

\section{Background and Related Works}\label{sec:back}

\subsection{Negation in Logic and Natural Language}

Negation is a complex phenomenon in both logics and natural language \cite{HW20}. In formal logics, the meaning of
negation seems to be easy: It is a unary operation toggling the truth value of
statements between true and false. In most classical logics, we have that double
negation does not alter the original truth value, whereas this might not be true
in other logics, e.g., multi-valued logics.
In any case, the scope of a negation is the subformula that is affected by the negation.
There are several types of negation in special logics: For instance, negation as
failure or default negation is a non-monotonic inference rule in logic
programming, used to derive $\mathit{not}~p$ for some statement $p$ from the failure to
derive $p$ provided that this is consistent with the current knowledge
\cite{Ant99}.\opt{long}{ In argumentation logics, strict and defeasible negation is
distinguished \cite{SL92,WS16}. The latter may be debated in the course of an
argumentation process until an argument is defeated or ultimately justified.}

In natural language, the meaning of negation often is not that clear, and it
consists of several components (cf. \cite{JM+20}): Instead of precisely defined
negation operators, there is a \emph{negation cue}, that may be a syntactically
independent negation marker like \enquote{never}, \enquote{nor}, \enquote{not}
(syntactic negation), an affix expressing negation like \enquote{i(n)-} and
\enquote{un-} (morphological negation), or a word whose meaning has a negative
component like \enquote{deny} (lexical negation). The \emph{scope} of a negation is the whole
part of the sentence or utterance that is negated, i.e., affected by the
negation cue.

Often the effect of negation shall be localized more precisely and reduced to
only one part of the scope. On the one hand, there is the negated event or
property, which we call \emph{negatus} henceforth, usually a verb, noun, or
adjective, that is directly negated. On the other hand, the \emph{focus} is the
part of the scope that is most prominently negated. It can also be defined as
the part of the scope that is intended to be interpreted as false or whose
intensity is modified \cite{JM+20}. By this, eventually a positive meaning
may be implied. For example, the sentence \enquote{They didn’t release the UFO
files until 2008.} has the negation cue \enquote{-n’t}. The pragmatic meaning of
the sentence is probably a positive one, namely \enquote{The UFO files were
released in 2008.}, since \enquote{until 2008} is the negation focus
\cite{morante-daelemans-2012-conandoyle}. Nevertheless, the negatus is
\enquote{release} inducing the temporal negative meaning of the sentence:
\enquote{They did not release the files before 2008}.

\subsection{Commonsense Reasoning and Negation}\label{sec:nl}

For automated commonsense reasoning, negation in natural language has to be
treated in a formal manner. Traditional approaches tackle this problem by using
Montague semantics together with Kripke semantics for modal logics \cite{JZ21}.
Here, negation often is discussed in the context of performative verbs, e.g.,
\enquote{promise}. So the two negations in \enquote{I do not promise not to assassinate the
prime minister.}~\cite{Lyo77} refer to different parts of the sentence and thus
do not annihilate each other like double negation in classical logics.
A more recent approach for formal negation detection comes from discourse
representation theory and categorial grammars \cite{KR93}. By means of the NLP
toolchain \cite{basile-etal-2012-ugroningen}, negation is detected on the basis
of this theory. By means of further tools like KnEWS~\cite{knews}, complete sentences possibly including negation can be
transformed eventually into a first-order logic formalization. Nevertheless, all
these procedures and tools cannot hide the fact that the meaning of negation in
natural language may be highly ambiguous. In languages like French or German,
e.g., even the scope of negation may not be clear because of syntactic
ambiguity. The German utterance \enquote{Ich verspreche dir nicht ruhig zu
bleiben.} may mean \enquote{I do not promise to stay calm.} or \enquote{I promise not
to stay calm.}

As written in the introduction, the motivation to investigate negation is
to treat it adequately in the context of cognitive reasoning. For this, it is
important to identify a negation by its cue and assign the corresponding negatus
correctly. For evaluation of our approach, there are
numerous benchmark sets for commonsense reasoning and negation: The \emph{COPA
Challenge} (Choice of Plausible Alternatives) \cite{copa} or the
\emph{StoryClozeTest} \cite{storyclozetest} require causal reasoning in everyday
situations, both without special focus on negation, however.
In \cite{morante-daelemans-2012-conandoyle}, \emph{ConanDoyle-neg}, a
corpus of stories by Conan Doyle annotated with negation information is
presented. Here, the negation cues and their scope as well as the negatus
have been annotated by two annotators in the CD-SCO dataset. The \emph{SFU Opinion and
Comments Corpus} \cite{KW+20} is a corpus for the analysis of online news
comments. One of its focuses is how negation affects the interpretation of
evaluative meaning in discourse.

\section{Methods}\label{sec:meth}

\subsection{A System for Cognitive Reasoning}\label{sec:corg}

The project CoRg (Cognitive Reasoning) \cite{SSS19a} aims at solving
commonsense reasoning benchmark tasks using extensive knowledge bases, automated
reasoning, and neural networks. Benchmarks for commonsense reasoning often are
multiple-choice natural language question-answering tasks. There, a premise
together with answer alternatives is given, and the objective is to find the
right alternative. An example for such a task is given in \cref{cloze}. It is
taken from the \emph{StoryClozeTest} benchmark \cite{storyclozetest}, which
contains 3,744 problems in total.

\begin{figure}
\textbf{Premise:} I think yesterday was when I realized I shouldn't be a
mathematician. I was at the grocery store, restocking for next week.
I had bought around twenty items, and thought I was under my budget.
When I went to check out, I was ten dollars over!

\textbf{Alternative 1:} I then realized that I was not good at math.\\
\textbf{Alternative 2:} Then I knew that I was great at math!
\caption{A StoryClozeTest example.}
\label{cloze}
\end{figure}

\begin{figure}\centering
  \includegraphics[width=0.5\textwidth]{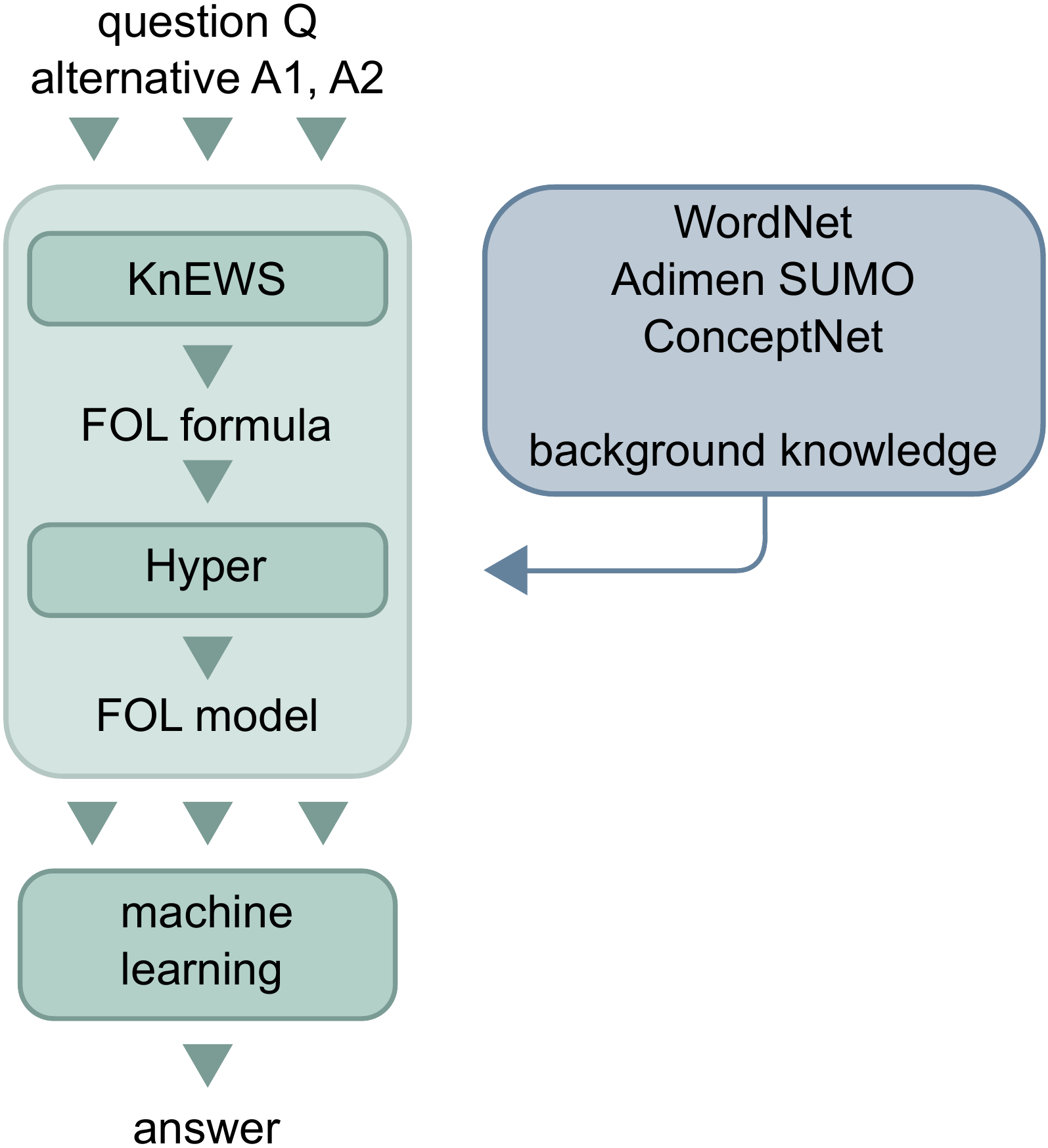}
  \caption{The CoRg system.}
  \label{corg}
\end{figure}

Most systems tackling this type of problems rely on pre-trained neural networks
and fine-tune their network for their specific problem. The natural language
input is mapped to word embedding vectors (cf. \cite{JF20}) and used as input for the
neural network. In our approach, however, we enrich the input by knowledge
derived through automated reasoning. The whole process is shown in \cref{corg}:
First, we translate
the natural language input into a logical formula using KnEWS \cite{knews}. This
is done for the premise as well as the answer candidates. Thus,
we generate three logical formulae.

In these formulae, the predicate names often correspond to words from the
original sentence. Thus, the predicate names are used to look up synonyms from WordNet
\cite{DBLP:journals/cacm/Miller95} and logically formalized background knowledge from
different ontologies like AdimenSumo
\cite{DBLP:journals/ijswis/AlvezLR12} and knowledge graphs like ConceptNet
\cite{Liu:2004:CMP:1031314.1031373}. For each of the three formulae generated by KnEWS, the gathered knowledge together with the
formula is used as input for the tableaux-based automated reasoning system Hyper
\cite{krhyper} to perform inferences. In this process, we do not expect Hyper to
find a proof, but use it as an inference engine to derive new knowledge. Hyper
thus computes a model or is terminated after a certain time and returns a
tableau branch computed up to that point. Since there is the possibility that
Hyper would still have closed this branch if it had run longer, we refer to this
branch as a \emph{potential partial model}. The information in the (potential
partial) models represents knowledge that can be (potentially) deduced from the
original sentence and the gathered background knowledge and thus provides more
information than the original natural language sentence alone.

We extract all predicate names from the output of Hyper and use them as input for a two-channel neural
network, pairing each answer candidate with the premise. Each word is mapped to
a word embedding vector. Here we make use of the 300-dimensional word embeddings
from ConceptNet Numberbatch \cite{speer2017conceptnet} which has 400,000
entries. The neural network calculates a likelihood how much the pairs (consisting of question and answer candidate)
correlate. Eventually, the answer with the highest likelihood is chosen as the
correct answer. For further details about the system, the reader is referred to
\cite{SSS19a,SSS19b}.

\subsection{Negation Scope and the Negatus -- Why Size Matters}
\label{sec:sizematters}

An unnecessarily large scope of a negation in a formula induces a problem for
the CoRg system. To see this, consider the following example (cf. first answer
alternative in \cref{cloze}):
\begin{quote}
\enquote{I then realized that I was not good at math.}
\end{quote}

KnEWS translates this sentence into the following formula:
\begin{align}
\exists A, B, C, D, E \Bigl(&person(A) \land person(B) \land \mathit{then}(C)\land manner(D,C)\land   \nonumber \\
 &\mathit{topic}(D,E) \land actor(D,B)\land realize(D)\land   \nonumber \\
&\neg  \exists F, G\bigl(at(G,F) \land math(F)\land \mathit{theme}(G,A)\land good(G)\bigr)\Bigr)  \label{eq:knewsformula_math}
\end{align}

We note that \cref{eq:knewsformula_math}, generated by KnEWS, is far from complete and how humans
would formalize this sentence. For instance, the generated formula introduces
two variables $A$ and $B$, which refer to the two occurrences of the personal
pronoun \enquote{I}, but sound anaphora resolution would yield $A=B$.
Furthermore, there is the singleton variable $E$, which actually refers to
the subclause \enquote{that I was not good at math}, but this is not formalized.
Nevertheless, \cref{eq:knewsformula_math} is still sufficient for our purposes.
Developing a more suitable translation from natural language to logic would
certainly be interesting\opt{long}{ (cf. \cite{BB+01})} but is beyond the scope
of our project.

The subformula affected by the negation obviously includes the predicates \emph{at},
\emph{math}, \emph{theme} and \emph{good}. It is easy to see that this scope is
unnecessarily large, since the word \enquote{math} is clearly not
negated in the sentence. Actually, it would be sufficient to only negate the
word \enquote{good}. Nevertheless, \enquote{not good} may have a different meaning
than \enquote{bad} for humans. This could be investigated by direct experiments
with humans, but this again is beyond the scope of our project. The big
negation scope is not only unpleasant from a theoretical point of view, but
leads to major problems in the CoRg system when the generated formula is fed
into the automated reasoning system Hyper.

Before we take a closer look at these problems, let us assume that the selected
background knowledge contains the information that math is a school subject. In
fact, ConceptNet contains the triple $(\mathit{Math, IsA, school\ subject)}$
which can be translated into:
\begin{equation}
\forall X \big(math(X) \rightarrow \mathit{school\_subject}(X)\big)\label{equ:formula_bg}
\end{equation}
In the CoRg system, this background knowledge together with
\cref{eq:knewsformula_math} is passed to the Hyper theorem prover. In
a preprocessing step, Hyper converts all input formulae to clause normal form.
For \cref{eq:knewsformula_math,equ:formula_bg}, this
includes the following clauses:
\begin{align}
\opt{long}{&person(a).\nonumber\\
&person(b).\nonumber\\
&then(c).\nonumber\\
&manner(d,c).\nonumber\\
&\mathit{topic}(d,e).\nonumber\\
&actor(d,b).\nonumber\\
&realize(d).\nonumber\\}
&at(G,F) \land math(F) \land theme(G,A) \land good(G)\rightarrow \bot. \label{eq:close_math}\\
&math(X) \rightarrow \mathit{school\_subject(X).} \label{eq:clause_bg}
\end{align}

Since Hyper is based on the hypertableau calculus
\cite{Baumgartner:Furbach:Pelzer:HyperTableauxCalculusEquality:2008}, it can use
\cref{eq:close_math} only to close branches. More precisely, forward reasoning
performed by Hyper is quite capable of using negative knowledge during the
reasoning process, but the inferred knowledge always contains only positive
statements. This means that the predicate name \emph{math} can never appear in
an open branch of a tableau constructed by Hyper. Moreover, \cref{eq:clause_bg},
created from the background knowledge, can not be used to infer that the person
in question is not good at a school subject. Therefore, the predicate names
\emph{math} and \emph{school\_subject} will not appear in (potential partial)
models or inferences of Hyper. As we have seen above, the machine learning
component of the CoRg system uses predicate names that occur in the models and
inferences of Hyper to decide which of the given alternatives provides the right
answer. Therefore, for us it is important that the output of Hyper contains as much
knowledge as possible from the natural language input together with inferred
knowledge from background knowledge.

In summary, the unnecessarily large scope of the negation in
\cref{eq:knewsformula_math} results in the fact that important
information is withheld from the machine learning component of the CoRg system.
Not all types of negation lead to this problem and need to be addressed by our approach: KnEWS is able to handle
lexical and morphological negations, as they are directly translated into a
single predicate without logical negation. Therefore, our approach only needs to
address syntactic negation like \enquote{not good}.
But syntactic negation is common in natural language: 127 of the first 310
examples of the StoryClozeTest \cite{storyclozetest} contain syntactic negation.

Since the way KnEWS handles syntactic negation leads to the above
mentioned problems in cognitive reasoning, we aim at determining the negatus for
a given negation and to reduce the scope of the negation in the formula to
include only the negatus. In the example, the subformula $good(G)$ represents
the desired negatus \enquote{good}. The reduction of the scope of the negation
in \cref{eq:knewsformula_math} to the negatus leads to the following formula:
\begin{align}
\exists A, B, C, D, E \Bigl(&person(A) \land person(B) \land then(C)\land manner(D,C)\land   \nonumber \\
 &topic(D,E) \land actor(D,B)\land realize(D)\land   \nonumber \\
&  \exists F, G(at(G,F) \land math(F)\land theme(G,A)\land \lnot good(G))\Bigr)  \label{eq:knewsformula_math_red}
\end{align}

We are aware that moving the negation in the formula to the predicate
representing the negatus does not preserve logical equivalence. But in our
practical application context of commonsense reasoning, this is not mandatory.
In addition, ordinary equivalence transformations such as De Morgan's rule do
not help to restrict the scope of the negation in the formula to the negatus.
But after reducing the scope of the negation of the formula to the negatus, in
many cases it is even possible to remove the negation completely by determining
an \emph{inverse} (antonym) for the predicate symbol of the negated subformula.
The negated subformula can be replaced by a non-negated subformula with the
inverse as predicate symbol.
For the example above, we determine the inverse $bad$ of the predicate symbol
$good$ of the negated subformula $\lnot good(G)$ with the help of WordNet and
substitute the negated subformula $\lnot good(G)$ in \cref{eq:knewsformula_math_red} by $bad(G)$.

Clausification of the resulting formula leads to a set of clauses consisting only of
unit clauses. Hence Hyper can use the background knowledge and infer that
\emph{math} is a \emph{school\_subject}. All this information will become
available to the machine learning component of the CoRg system.
Of course, one could argue that \enquote{being not good at math} is not the same
as \enquote{being bad at math}, but since our goal is to allow the theorem
prover to draw as many inferences as possible, we accept this imprecision and
consider the replacement by an antonym as an approximation.

\subsection{Approach to Negation Treatment for Cognitive Reasoning}\label{sec:treatement}

Now, after having described our motivation for the determination of the exact
negatus, we describe the implementation and evaluation of this approach.
For a given text $T$ and its corresponding first-order logic representation $F$,
\cref{alg:reduction} describes how we treat negations. In the beginning, some
preprocessing is performed (\cref{line:1,line:2}): For the formula $F$, this means
the removal of double negations. For the text $T$, this step includes lemmatization,
tokenization, part-of-speech (POS) tagging and removal of certain stopwords like
\enquote{the} or \enquote{he}.
\begin{algorithm}
\LinesNumbered
\SetAlgoLined
\caption{Treat negations}\label{alg:reduction}
\KwIn{\\
	$T$: natural language text (input from benchmark sample)\\
	$F$: first-order logic formula for $T$ (output from KnEWS)\\
	$k$: size of word window\\
	$\mathit{strategy}$: to determine the negatus}
  \smallskip
\KwResult{\\
	$A$: assignments $x_i \leftrightarrow y_j$ between negation cues and logical negations\\
	$N$: negatus for each assignment $A$}
  \smallskip
	preprocess formula $F$ (remove double negations)\label{line:1}\\
	preprocess text $T$ (lemmatization, tokenization, POS tagging, removal of stopwords)\label{line:2}\\
  \smallskip
	$x_1,\dots,x_n$ = negation cues in text $T$\label{line:3}\\
	\For{$i=1,\dots,n$}{$M(x_i)$ = word window (set) of size $k$ around negation cue $x_i$ (depends on $\mathit{strategy}$)}
	$y_1,\dots,y_m$ = logical negations in first-order formula $F$\\
	\For{$j=1,\dots,m$}{$M(y_j)$ = set of predicates occurring in the scope of negation $y_j$}\label{line:4}
  \smallskip
	\For{$i=1,\dots,n$\label{line:5}}{
	$j = \mathrm{argmax}_\ell \big\{ |M(x_i) \cap M(y_\ell)| : \ell \in \{1,\dots,m\} \big\}$\\
	$i' = \mathrm{argmax}_\ell \big\{ |M(x_\ell) \cap M(y_j)| : \ell \in \{1,\dots,n\} \big\}$\\
	\If{$i'=i\;\wedge\;M(x_i) \cap M(y_j) \neq \emptyset$}{
		$A_{ij}$ = assignment $x_i \leftrightarrow y_j$\\
		$N_{ij}$ = word $w$ in $M(x_i)$ (negatus) according to $\mathit{strategy}$
	}}\label{line:6}
\end{algorithm}
\opt{long}{The stopwords we use are based on the stopwords used in the Natural
Language Toolkit (NLTK) Python library for Natural Language
Processing.\footnote{Available at: \url{http://www.nltk.org/}} Since modal and
auxiliary verbs and negation cues are important for our purposes, we do not take
them as stopwords. \cref{stopwords} shows the complete set of used stopwords.

\begin{figure}
\begin{quote}\small
a,
about,
above,
after,
again,
against,
ain,
all,
am,
an,
and,
any,
are,
as,
at,
because,
been,
before,
being,
below,
between,
both,
but,
by,
d,
did,
do,
does,
doing,
down,
during,
each,
few,
for,
from,
further,
having,
he,
her,
here,
hers,
herself,
him,
himself,
his,
how,
i,
if,
in,
into,
it,
it's,
its,
itself,
just,
ll,
m,
ma,
me,
more,
most,
my,
myself,
now,
o,
of,
off,
on,
once,
only,
or,
other,
our,
ours,
ourselves,
out,
over,
own,
re,
s,
same,
shan,
she,
she's,
should've,
so,
some,
such,
than,
that,
that'll,
the,
their,
theirs,
them,
themselves,
then,
there,
these,
they,
this,
those,
through,
to,
too,
under,
until,
up,
ve,
very,
we,
what,
when,
where,
which,
while,
who,
whom,
why,
will,
with,
won,
y,
you,
you'd,
you'll,
you're,
you've,
your,
yours,
yourself,
yourselves
\end{quote}
\caption{Complete set of used stopwords.
Note that all words in this list are lower case. So the word \enquote{i} in this
list corresponds to \enquote{I}. Furthermore, single letters (like \enquote{s})
in this list are included in the stopwords since they can be used to remove
parts of speech resulting from tokenization. For instance, \enquote{he's} is
tokenized into \enquote{he} and \enquote{s}.}
\label{stopwords}
\end{figure}}

Then (in \crefrange{line:3}{line:4}) the negation cues $x_1,\dots,x_n$ occurring
in the text $T$ as well as the logical negations $y_1,\dots,y_m$ in the formula
$F$ are considered. For each negation cue $x_i$, the set $M(x_i)$ of all words
in a word window of size $k$ around (usually after) the negation cue are
determined. In our evaluation we use $k =3$ which has been established
by manually determining the position of the negatus relative to the position
of the negation cue from 486 example negations of the CD-SCO train dataset\opt{long}{
(cf. \cref{other.fig} for details)}.
Other choices are possible and will be considered in
future work. Depending on the strategy we consider the word preceding the negation cue to the word window.
Similarly, for each logical negation $y_j$, the set $M(y_j)$ of
predicates occurring in the scope of negation $y_j$ is determined.

\opt{long}{\begin{figure}[ht]
\includegraphics[width=0.9\textwidth]{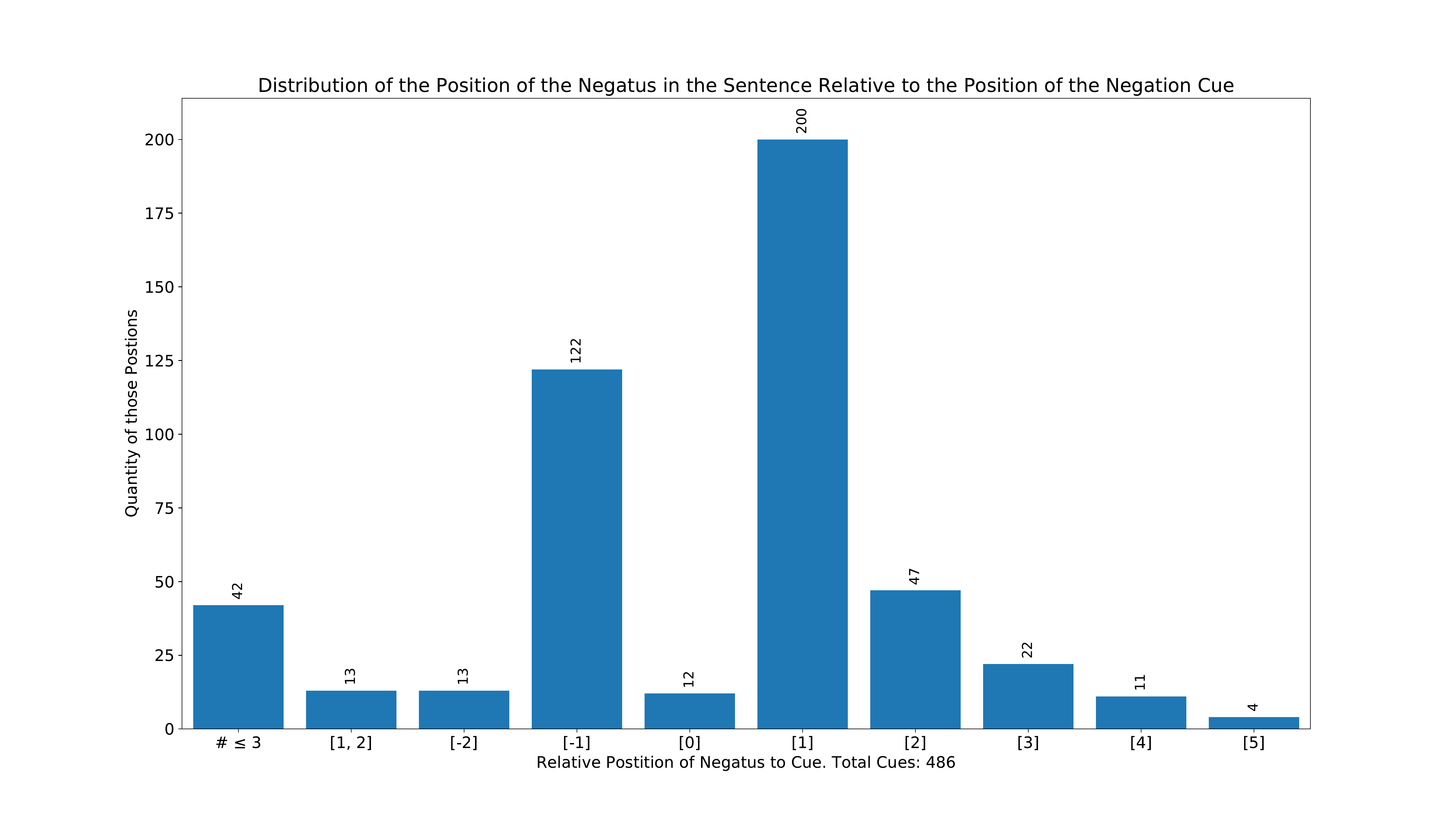}
\caption{Distribution of the relative position of the negatus in the sentence.
Positive position values indicate that the negatus occurred after the negation
cue, negative values that it occurred before the cue in the sentence. For
instance, in 200 cases the negatus was the word immediately following the
negation cue (indicated by [1]) and in 122 cases the negatus was the word
immediately preceding the cue. The position cue [1,2] in the $x$-axis denotes a
negatus consisting of two words, the first in position~1 and the second in
position~2 after the negation cue. The bar annotated with $\# \leq 3$ comprises
negatus positions which have less or equal $3$ occurrences.}
\label{other.fig}
\end{figure}}

Texts may contain several negations. In general, the number $n$ of negation cues
in the text does not coincide with the number $m$ of logical negations in the
formula. Therefore, it can happen that for some negation in the text no
corresponding negation can be found in the formula or vice versa. Since in
addition the order of the information in the text does not always coincide with
the order of the information in the generated formula, the first negation in the
text does not necessarily correspond to the first negation in the formula.
Hence, it is necessary to determine a mapping between both kinds of
negation. We do this as follows: The negation cue $x_i$ and the logical negation
$y_j$ are assigned to each other if the intersection of $M(x_i)$ and $M(y_j)$ is
non-empty and has maximal cardinality with respect to both indexes $i$ and $j$
(cf. \crefrange{line:5}{line:6}).

Finally, for each assignment $x_i \leftrightarrow y_j$, the negatus $N_{ij}$ is
determined. Which word is selected as negatus, depends on the specified strategy
and thus on the word types of the words in the word window, including possibly
the negation cue. The strategies considered are:
\begin{itemize}
\item \emph{baseline:} For a negation cue that is a negated
modal verb (\enquote{can't}, \enquote{couldn't}, \enquote{should not}), the modal
verb is chosen as the negatus (i.e., \enquote{can}, \enquote{could}, \enquote{should}).
For all other cases, the negatus is the first non-stopword in the word window. The baseline strategy was determined after manually
examining the position of the negatus relative to the position of the negation
cue\opt{long}{ (cf. \cref{other.fig})}.
\item \emph{first non-stopword strategy (FNS):} Negatus is the first non-stopword in the word window.
\item \emph{first verb strategy (FV):} Negatus is the first verb in the word window.
\item \emph{first verb or non-stopword strategy (FNS+FV):} Negatus is the first verb in the word window.
Use \emph{first non-stopword strategy} if no negatus was determined this way.
\item \emph{combination strategy (Comb):} Determine negatus word according to \cref{tab:components}.
\end{itemize}
All strategies use a word window consisting of the first $k$ non-stopwords
following the negation cue -- with one exception: The combination strategy
uses a word window of size $k$ which starts from the first non-stopword left of
the negation cue. The negation cue is not included in the word window.
Afterwards, if the determined negatus corresponds to a predicate name in the
scope of the negation, formula $F$ is adapted such that the scope only includes
this predicate. Further, if possible, the negatus predicate is replaced by its
inverse, i.e., an antonym, such that eventually the negation is completely
removed. Antonyms are looked up in WordNet.

\begin{table}
\begin{center}
\caption{Procedure to determine the negatus in the \emph{combination strategy}
for syntactic negation. For each negation cue, the table is scanned from top to
bottom and the first row matching the cue is selected and applied.}\label{tab:components}
\begin{tabular}{ll} \toprule
\textbf{Negation Cue}& \textbf{Negatus}\\
 \midrule
can't, cannot, can not & can\\
couldn't, could not		& could\\
shouldn't, should not		& should\\
nothing, isn't, is not, aren't, are not, wasn't, was not,	\phantom{bla}	& \\%first non-stopword in word window\\
weren't, were not & first non-stopword in word window\\
no					& first noun after negation\\
not, never, all other negation cues 		& first verb in word window\\
any cues, if no negatus has been determined yet 		& first non-stopword in word window\\
 \bottomrule
 \end{tabular}
 \end{center}
\end{table}

\opt{long}{\subsection{Example for \cref{alg:reduction}}

To further illustrate this procedure, we consider the following example:
\begin{quote}
\enquote{I \textcolor{red}{don't like} the \textcolor{red}{cookies} with
\textcolor{red}{raisins} but I \textcolor{cyan}{can't eat} the
\textcolor{cyan}{ones} with \textcolor{cyan}{chocolate} either.}
\end{quote}
For convenience, the important parts are colored. The corresponding formula is:
\begin{align*}
\exists A, B, C, D \, \Bigl( one(A) \land person(C) \land cookie(B) \land person(D) \land \nonumber\\
\neg \exists E, F, G, H, I \, \bigl( either(G) \land manner(I,G) \land with(I,H) \land chocolate(H) \nonumber \land \\
theme(I,A) \land actor(I,C) \land eat(I) \land  topic(F,E) \land actor(F,C) \land  can(F) \land \nonumber \\
\neg  \exists J, K \, ( theme(J,B) \land  actor(J,D) \land like(J) \land  with(B,K) \land  raisin(K))\bigr)\Bigr) \label{eq:knewsformula_example_alg}
\end{align*}
The cues and word windows are as follows, with $x_i$ denoting the words of the
natural language representation and $y_i$ denoting the parts of the formula:
\begin{align}
&x_1 = \text{don't} &&M(x_1) =\{\text{like, cookies, raisins}\} \nonumber\\
&x_2 = \text{can't} &&M(x_2) =\{\text{eat, ones, chocolate}\} \nonumber\\
&y_1 = \text{not} &&M(y_1) =\{\text{either, with, chocolate, eat, can}\} \nonumber\\
&y_2 = \text{not} &&M(y_2) =\{\text{like, raisin}\} \nonumber
\end{align}
One can see that the negations are in reverse order here. Calculating the
cardinality of the intersections for every combination of $M$ (after reducing
all words to their base form, e.g., \enquote{raisins} to \enquote{raisin})
leads to the following:
\begin{align*}
&|M(x_1) \cap M(y_1)| = 0 &&|M(x_2) \cap M(y_1)| = 2 \nonumber\\
&|M(x_1) \cap M(y_2)| = 2 &&|M(x_2) \cap M(y_2)| = 0 \nonumber
\end{align*}

This leads to the assignments $x_1 \leftrightarrow y_2$ as well as $x_2 \leftrightarrow y_1$.
To determine the negatus, we apply, e.g., the FNS-Strategy to the natural language text,
which results in $\mathit{negatus}_1=$~\enquote{like} and
$\mathit{negatus}_2=$~\enquote{can}. Their respective inverse words, which can
be used to replace the negated predicates in the formula, are $\mathit{inverse}_1
=$~\enquote{dislike} and $\mathit{inverse}_2=$~\enquote{unable}.}

\section{Experiments}\label{sec:eval}

To evaluate our approach, we use the *SEM 2012 Negation Task
\cite{morante-blanco-2012-sem,morante-daelemans-2012-conandoyle} as benchmark,
where negations in some books by Conan Doyle are annotated (cf.
\cref{sec:nl}), more precisely CD-SCO, the corpus with scope annotation that
additionally provides a negatus. The sentences of the corpus are split
into words which are POS-tagged. Furthermore the stem of the word and a
syntax-tree for the sentence is given.

However, the *SEM 2012 Negation Task does not entirely fit to our problem. Since we only want
to address syntactic negation (cf. \cref{sec:sizematters}), we manually created
a filtered version of CD-SCO: Lexical and morphological negations are
deleted, while the syntactic ones remain. This process resulted in 132 deleted
samples in the train dataset, 32 deleted samples in the development (dev)
dataset and 32 deleted samples in the test dataset. Additionally, we manually
created an own benchmark based on the StoryClozeTest \cite{storyclozetest},
taking the first 100 tasks of the StoryClozeTest that contain a syntactic
negation. Each StoryClozeTest task consists of six sentences. We generated one
negation task for each negation we found (126) and the second and third author
of this paper annotated the negatus for each negation. We denote these
benchmarks as \emph{Cloze-NEG}.
In addition to that, we evaluated our approach within the CoRg system using all
problems from the StoryClozeTest.\footnote{All used benchmark sets\opt{short}{,}
\opt{long}{and }our implementation in Python\opt{short}{, and an extended
version of this paper} are available at \opt{short}{\url{http://arxiv.org/abs/2012.12641}}%
\opt{long}{\url{http://gitlab.uni-koblenz.de/obermaie/negationcognitivereasoning}}.}

Next, we present experimental results of our approach on the CD-SCO benchmarks
and compare them with results achieved by different systems in the
*SEM 2012 negation task on scope resolution where also the negatus has been
determined. For the *SEM 2012 Shared Task, it was possible to enter systems in
two tracks: the closed and the open track. The systems in the closed track did not use any
external resources. In contrast to this, the systems in the open track were
allowed to use external resources as well as tools. Since our approach relies on
the KnEWS system which is built on top of Boxer \cite{bos-2008-wide}, our
approach belongs to the open track and we thus compare our
approach to the systems from the open track in the following.
Furthermore, we use evaluation measure $B$
of the *SEM 2012 competition where precision is computed as the number of true positives divided
by the total number of a systems predictions.

\opt{long}{\subsection{Other Approaches}

We compare our approach against other approaches, in particular the five
approaches submitted to the open track of the *SEM 2012 Negation Task
\cite{morante-blanco-2012-sem,morante-daelemans-2012-conandoyle}: UiO2,
UGroningen r2, UGroningen r1, UCM-1 and UCM-2.
UiO2 uses conditional random fields (CRFs) and support vector machines (SVMs),
while the UGroningen approaches use discourse representation structures (DRSs),
and the UCM approaches rely on syntax trees and rules. Let us discuss these
approaches in some more detail:

UiO2 \cite{lapponi2012uio} is the only approach using machine learning. An SVM
classifier detects cues, while the CRF classifiers \cite{lavergne2010practical}
predict the scopes and events. In the open track, UiO2 makes use of the MaltParser
\cite{nivre2006maltparser} and its pre-trained model to obtain dependency
graphs.

The UGroningen approaches \cite{basile-etal-2012-ugroningen} are probably the
most similar to our approach, as they use C\&C Tools and Boxer
\cite{bos-2008-wide,CurranClarkBos2007ACL} to produce DRSs. We use the tool
KnEWS, which integrates C\&C Tools and Boxer.
Another similarity is the detection of the scope, which is extracted from the
tokens that occur in the scope of the negated DRS.

The UCM approaches \cite{de2012ucm,ballesteros2012ucm} implement a rule-based
system. They identify the cues using a list of predefined negation words. For
lexical negation cues, they add a step using WordNet antonyms to make sure that
it is a negated form of an existing word. Scope detection is done by using
the syntax tree of the sentence. To identify the negatus they have two
strategies: If it is a lexical negation they mark the word itself as negatus, if
it is a syntactical negation they mark the verb as the negatus, if it is also the
next word. Our approach can be seen as a combination of the UGroningen and UCM
approaches.}

\subsection{Data Preparation and Evaluation}

\cref{tab:cd-sco} shows the
results of our approach using different strategies and results of systems
in the open track of *SEM 2012 Task~1. Since we do not use machine learning in
this procedure, we present our results for the training and development set as
well. For the systems of *SEM 2012 Task~1, only the results on the CD-SCO test
set are shown.
\begin{table}[t]
\caption{Results of our approach using the strategies described in
\cref{sec:treatement} on the CD-SCO dataset in the upper part of the table. The
lower part of the table depicts the result of the systems in the open track of
*SEM 2012 Task~1.}\label{tab:cd-sco}
\centering
\begin{tabular}{lccccccccccc}%cccc}
\toprule
& \multicolumn{3}{c}{\textbf{Train}} & \phantom{b}& \multicolumn{3}{c}{\textbf{Dev}} & \phantom{b}& \multicolumn{3}{c}{\textbf{Test}}\\%& \phantom{b}& \multicolumn{3}{c}{\textbf{Test Filtered}}\\
\cmidrule{1-12}
  & Prec. & Rec. & $F_1$ & \phantom{b}& Prec. & Rec. & $F_1$ & \phantom{b}& Prec. & Rec. & $F_1$ \\%& \phantom{b}& Prec. & Rec. & $F_1$\\
\midrule
Baseline  &24.56  &34.31 &28.63 & &26.17& 31.97&28.78  & &23.14 &32.37 &26.99 \\%&&25.00&.35.00&29.17\\
FNS         & 24.56 &34.31 &28.63 & &26.17 &31.98 &28.78 & &23.14&32.37&26.99\\%&&25.00&.35.00&29.17\\
FV           & 19.75 & 20.65& 20.19& &22.52 & 20.49&21.46 && 24.42&24.28 &24.35\\%&&26.57&27.14&26.86\\
FV+FNS & 22.47 & 31.38 & 26.19& &26.85&32.79 & 29.52& & 23.55&32.95&27.47\\%&&25.00&35.00&29.17\\
%FN &  & & & & & & & &&&\\
Comb. & 35.39 &49.43 &41.25 & & 34.90& 42.62&38.38 &&34.71&48.55 &40.48\\%&&32.14&45.00&37.50\\
\midrule
UiO2 &  & & & & & &&&63.82 & 57.40 &60.44\\%&&&&\\
UGroningen r2 &  & & & &&& & & 55.22& 65.29 & 59.83\\%&&&&\\
UGroningen r1 &  & & & &&& & &52.66 &52.05 &52.35\\%&&&&\\
UCM-1 &  & & & & & &&&66.67 &12.72 &21.36\\%&&&&\\
UCM-2 &  & & & & & &&&44.44 & 21.18&28.69\\%&&&&\\
\bottomrule
\end{tabular}
\end{table}
In the CD-SCO benchmarks, a significant proportion of examples with negation
have a cue and a scope in the gold standard, but no negatus.
So, the train dataset contains 3,643 text passages with a
total of 3,779 lines in the gold standard data. Of these lines, 983 contain a
negation. However, in only 615 of these cases, a negatus is specified in the gold
standard. In contrast to this, our approach often finds a negatus in these cases.
\opt{long}{For instance, consider example \emph{baskervilles07\_301} from the
train dataset which contains the sentence: \enquote{I must not stop, or my
brother may miss me.} The gold standard does not specify a negatus for this
example. The formula created by KnEWs is as follows:
\begin{align*}
\exists A,B,C,D \Bigl( person(A) \land brother(B) \land \mathit{of}(B,C) \land person(C) \land person(D) \land \nonumber\\
\lnot \exists E,F,G \bigl(actor(G,D) \land \mathit{stop}(G) \land \mathit{topic}(F,E) \land actor(F,D) \land must(F)) \lor\nonumber \\
\exists H,I,J (theme(J,A) \land actor(J,B) \land miss(J) \land \mathit{topic}(I,H) \land
\mathit{actor}(I,B) \land may(I)\bigr)\Bigr)\nonumber
\end{align*}

Although no negatus is specified in the gold standard for this example, we can remove the
negation in the formula with our approach by replacing the predicate \emph{stop}
(corresponding to the negatus) by its inverse, namely the antonym
\emph{proceed}. This result corresponds to the sentence \enquote{I must proceed,
or my brother may miss me.}.
Cases like this example are counted as false in the calculation of
precision in \cref{tab:cd-sco}. }
We report numbers for the CD-SCO benchmarks in \cref{tab:cd-sco_langstrumpf} for
the filtered CD-SCO benchmarks where only negations with negatus specification
in the gold standard are considered.
Of course, sometimes there are cases where indeed no negatus should
be specified like in direct speech and elliptic utterances, where negations refer
back to the previous sentence. \opt{long}{Additionally sometimes the gold
standard contain errors. One example is from the train set with a wrongly
annotated negatus: \enquote{The dinner itself was neither well served nor well
cooked [\dots]}. The gold standard states \enquote{served} and \enquote{cooked} as
negatus, while we are confident that the negatus in both cases should be
\enquote{well}.}

\begin{table}[t]
\centering
\caption{Results of our approach using the strategies described in
\cref{sec:treatement} on the filtered CD-SCO dataset. For the calculation of
precision cases where our approach found a negatus but the gold standard
did not specify a negatus were not included.}\label{tab:cd-sco_langstrumpf}
\begin{tabular}{lccccccccccc}
\toprule
& \multicolumn{3}{c}{\textbf{Train Filtered}} & \phantom{b}& \multicolumn{3}{c}{\textbf{Dev Filtered}} & \phantom{b}& \multicolumn{3}{c}{\textbf{Test Filtered}}\\
\cmidrule{1-12}
  & Prec. & Rec. & $F_1$ & \phantom{b}& Prec. & Rec. & $F_1$ & \phantom{b}& Prec. & Rec. & $F_1$ \\
\midrule
Baseline &44.56&43.83&44.19 &&44.09&44.09&44.09 &&42.25&41.10&41.67\\
FNS       &44.56&43.38&44.19 &&44.09&44.09&44.09 &&42.25&41.10&41.67\\
FV         &36.71&26.13&30.53&&42.86&29.03&34.62 &&44.21&28.77&34.85\\
FV+FNS& 40.79&40.12&40.46&&47.31&47.31&47.31&&42.96&41.78&42.36\\
%FN &  & & & & & & & &&&\\
Comb.   &64.02& 62.96& 63.49&&58.06 &58.06 &58.06 &&61.97 &60.27 &61.11\\
\bottomrule
\end{tabular}
\end{table}

Although the CD-SCO benchmarks do not exactly reflect what we are trying to
achieve with our approach, the option combination strategy still ranks in the
midfield of the systems that participated in *SEM 2012.
In addition to that, we tested our approach on the 100 manually labeled StoryClozeTest
tasks, the Cloze-NEG benchmarks. The results are depicted in
\cref{tab:roc}.
To check whether our approach is suitable for our purpose of identifying the
negatus, the Cloze-NEG benchmarks are the most suitable ones. This is because
only these
benchmarks aim to determine a word in the scope of the negation that, when
replaced by its antonym, leads to the removal of the negation. The results on
the StoryClozeTest benchmarks show that the option combination strategy is often
able to identify the negatus.
Furthermore, we observe that for all considered benchmarks, the option
combination strategy seems to be the most suitable for identifying the negatus.

\begin{table}\centering\vspace*{-5pt}
\caption{Results of our approach on the 100 manually labeled StoryClozeTest tasks.}\label{tab:roc}
\vspace*{3mm}
\begin{tabular}{lccc}
\toprule
& \multicolumn{3}{c}{\textbf{Cloze-NEG}} \\
\cmidrule{1-4}
  & Prec. & Rec. & $F_1$ \\
\midrule
Baseline &59.52&59.52&59.52 \\
FNS &57.85&55.56&56.68 \\
FV &54.33&49.21&51.67\\
FV+FNS &56.35 &56.35&56.35\\
%FN &  & & & & & & & &&&\\
Comb. & 66.67 &66.67&66.67\\
\bottomrule
\end{tabular}
\end{table}

As explained above, an unnecessarily large scope of a negation in a formula
belonging to a commonsense reasoning problem causes the Hyper reasoner in the
CoRg system to perform only few inferences and, in the extreme case, to return
an empty model. To evaluate the benefit of our approach within the CoRg system,
we integrated our approach into the system and used it on the whole set of
problems in the StoryClozeTest.
By this, the amount of completely empty models was reduced by 71.95\% from 164
to 46\opt{long}{ (cf. \cref{fig:model_word_count})}. Overall, the negation
treatment resulted in an increase in the average number of distinct predicate
names per model from 18.83 to 19.35 with an overall gain of distinct predicate
names of 5,726 in 11,119 models. This demonstrates that our negation handling
enables the theorem prover to derive additional information, which in return can
be used by further processing, in our case the neural network. We hope that
these encouraging results will facilitate using automated reasoning in
commonsense reasoning.

\opt{long}{\begin{figure}[ht]
\includegraphics[width=\textwidth]{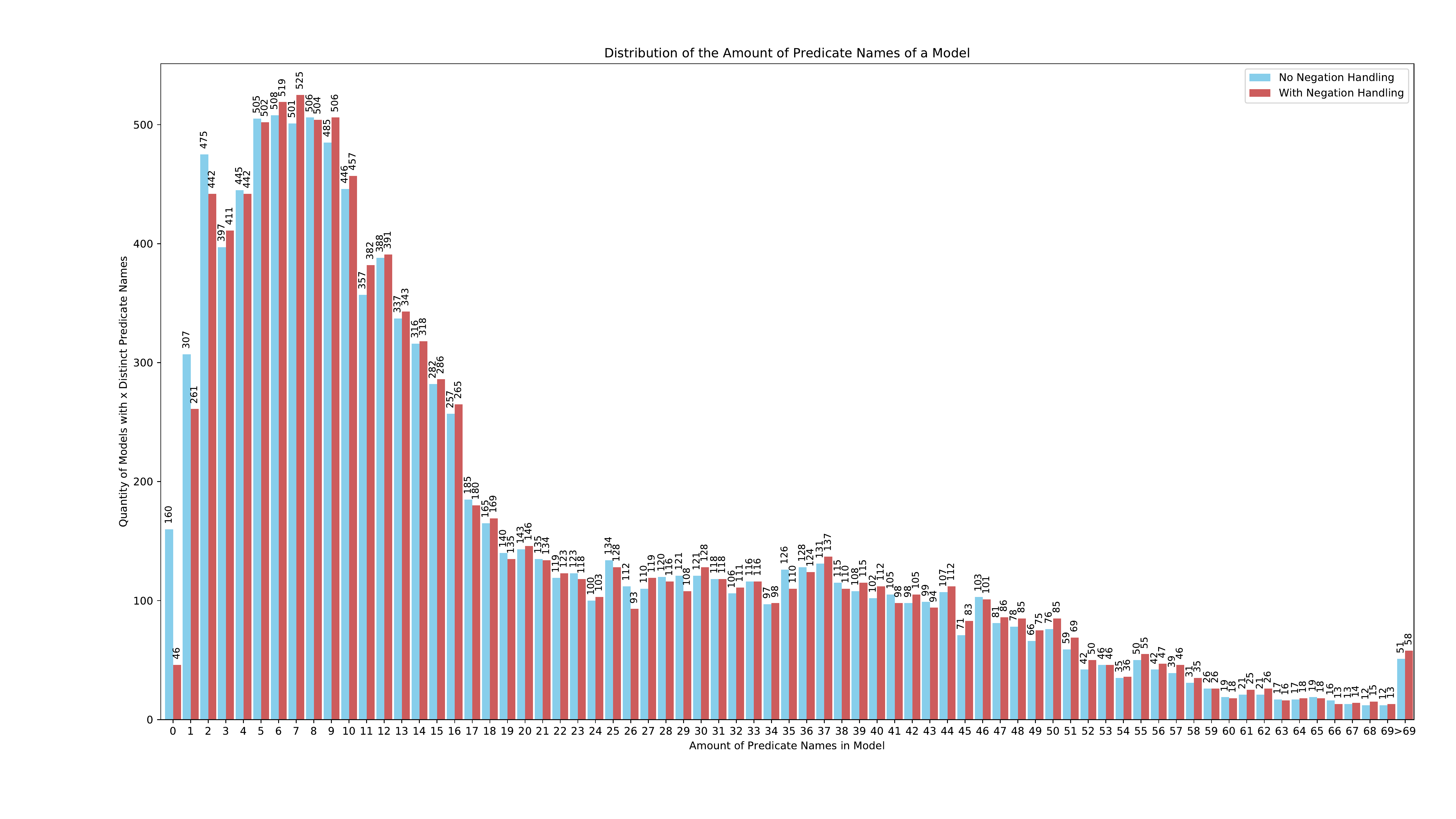}
\caption{Distribution of the number of distinct predicate names in a model generated by Hyper with
and without negation handling. WordNet was added as background knowledge. Models
with more than 69 elements are comprised in one entry to enhance readability. The empty models where reduced
from 164 to 46, which is an decrease of 71.95\%. The average number of distinct
predicates per model increased from 18.83 to 19.35, while the absolute amount of
distinct predicates increased from 209,374 to 215,100 with 5,726 new distinct
predicates. Overall 11,119 models where examined. One can see that not only the
elimination of empty models lead to the increase, but other models benefit from
our procedure as well. For instance, the amount of models containing only one
predicate decreased from 307 to 261.}
\label{fig:model_word_count}
\end{figure}}

\opt{long}{\vspace*{-1pt}}
\section{Summary, Conclusions, and Future Work}\label{sec:final}

In this paper, we presented a way to detect syntactic negation in natural
language using KnEWS, which is important in cognitive reasoning systems. This is
why we want to replace negative information through a positive counterpart, its
inverse. To achieve this, the negatus has to be identified and later on be
replaced by its inverse. Our approach detects syntactic negation and identifies
the negatus by combining the information from the generated KnEWS formula and
from the natural language input itself.

We apply different strategies for identifying the
negatus. The combination strategy apparently works best and ranks in the midfield
of the *SEM 2012 negation task. Our evaluation of the StoryClozeTest yields
comparable results. With the help of our approach for negation handling, we
improved the input of the theorem prover Hyper in the CoRg system. This resulted
in a decrease of the number of empty models by 71.95\% and an average increase
of 0.53 distinct predicates per model. This demonstrates an information gain
which is useful for later processing in the CoRg system.

In future work, we aim to improve our negation handling by applying other strategies in
more detail. Furthermore we shall use the inverse of the negatus in commonsense
reasoning. So far, we implemented a first approach looking up the antonyms of
the detected negatus from WordNet. Since eventually we work with neural
networks and word embeddings (cf. \cite{SSS19b}), another idea is to directly
generate word embeddings by using an inverse operator.\opt{long}{ The thus
determined antonym does not necessarily need a natural language counterpart, but
would be used as an artificial word.}
The overall goal is to implement a cognitive reasoning system with an adequate
treatment of syntactic negation.

\bibliographystyle{splncs04}
\bibliography{lit}

\begin{thebibliography}{10}
\providecommand{\url}[1]{\texttt{#1}}
\providecommand{\urlprefix}{URL }
\providecommand{\doi}[1]{https://doi.org/#1}

\bibitem{starsem2012}
Agirre, E., Bos, J., Diab, M., Manandhar, S., Marton, Y., Yuret, D. (eds.):
  *{SEM} 2012: The First Joint Conference on Lexical and Computational
  Semantics -- Volume 1: Proceedings of the main conference and the shared
  task, and Volume 2: Proceedings of the Sixth International Workshop on
  Semantic Evaluation ({S}em{E}val 2012). Association for Computational
  Linguistics, Montr{\'e}al, Canada (2012),
  \url{http://www.aclweb.org/anthology/volumes/S12-1/}

\bibitem{de2012ucm}
de~Albornoz, J.C., Plaza, L., D{\'\i}az, A., Ballesteros, M.: {UCM-1}: A
  rule-based syntactic approach for resolving the scope of negation. In: Agirre
  et~al.  \cite{starsem2012}, pp. 282--287,
  \url{http://www.aclweb.org/anthology/S12-1037}

\bibitem{DBLP:journals/ijswis/AlvezLR12}
{\'{A}}lvez, J., Lucio, P., Rigau, G.: {A}dimen-{SUMO}: Reengineering an
  ontology for first-order reasoning. Int. J. Semantic Web Inf. Syst.
  \textbf{8}(4),  80--116 (2012), \url{http://doi.org/10.4018/jswis.2012100105}

\bibitem{Ant99}
Antoniou, G.: A tutorial on default logics. ACM Computing Surveys
  \textbf{31}(4),  337--359 (1999), \url{http://doi.org/10.1145/344588.344602}

\bibitem{ballesteros2012ucm}
Ballesteros, M., D{\'\i}az, A., Francisco, V., Gerv{\'a}s, P., De~Albornoz,
  J.C., Plaza, L.: {UCM-2:} a rule-based approach to infer the scope of
  negation via dependency parsing. In: Agirre et~al.  \cite{starsem2012}, pp.
  288--293, \url{http://www.aclweb.org/anthology/S12-1038.pdf}

\bibitem{basile-etal-2012-ugroningen}
Basile, V., Bos, J., Evang, K., Venhuizen, N.: {UG}roningen: Negation detection
  with discourse representation structures. In: Agirre et~al.
  \cite{starsem2012}, pp. 301--309,
  \url{http://www.aclweb.org/anthology/S12-1040}

\bibitem{knews}
Basile, V., Cabrio, E., Schon, C.: {KNEWS}: Using logical and lexical semantics
  to extract knowledge from natural language. In: Proceedings of the European
  Conference on Artificial Intelligence (ECAI) (2016),
  \url{http://hal.inria.fr/hal-01389390}

\bibitem{Baumgartner:Furbach:Pelzer:HyperTableauxCalculusEquality:2008}
Baumgartner, P., Furbach, U., Pelzer, B.: The hyper tableaux calculus with
  equality and an application to finite model computation. Journal of Logic and
  Computation  \textbf{20}(1),  77--109 (2010),
  \url{http://doi.org/10.1093/logcom/exn061}

\bibitem{krhyper}
Bender, M., Pelzer, B., Schon, C.: System description: {E-KRHyper} 1.4 --
  extensions for unique names and description logic. In: Bonacina, M.P. (ed.)
  Automated Deduction -- {CADE}~24. pp. 126--134. LNCS~7898, Springer (2013),
  \url{http://doi.org/10.1007/978-3-642-38574-2_8}

\bibitem{BB+01}
Blackburn, P., Bos, J., Kohlhase, M., de~Nivelle, H.: Inference and
  computational semantics. In: Bunt, H., R., M., Thijsse (eds.) Computing
  Meaning. pp. 11--28. No.~77 in Studies in Linguistics and Philosophy,
  Springer, Dordrecht (2001),
  \url{http://link.springer.com/chapter/10.1007/978-94-010-0572-2_2}

\bibitem{bos-2008-wide}
Bos, J.: Wide-coverage semantic analysis with {B}oxer. In: Semantics in Text
  Processing. {STEP} 2008 Conference Proceedings. pp. 277--286. College
  Publications (2008), \url{http://www.aclweb.org/anthology/W08-2222}

\bibitem{CurranClarkBos2007ACL}
Curran, J.R., Clark, S., Bos, J.: Linguistically motivated large-scale {NLP}
  with {C\&C} and {B}oxer. In: Proceedings of the ACL 2007 Demo and Poster
  Sessions. pp. 33--36. Prague, Czech Republic (2007),
  \url{http://aclanthology.org/P07-2009/}

\bibitem{DBLP:conf/naacl/DevlinCLT19}
Devlin, J., Chang, M., Lee, K., Toutanova, K.: {BERT:} pre-training of deep
  bidirectional transformers for language understanding. In: Burstein, J.,
  Doran, C., Solorio, T. (eds.) Proceedings of the 2019 Conference of the North
  American Chapter of the Association for Computational Linguistics: Human
  Language Technologies, {NAACL-HLT}, Volume 1 (Long and Short Papers). pp.
  4171--4186. Association for Computational Linguistics (2019),
  \url{http://aclweb.org/anthology/papers/N/N19/N19-1423/}

\bibitem{FH+19a}
Furbach, U., H{\"o}lldobler, S., Ragni, M., Schon, C., Stolzenburg, F.:
  Cognitive reasoning: A personal view. KI  \textbf{33}(3),  209--217 (2019),
  \url{http://link.springer.com/article/10.1007/s13218-019-00603-3}

\bibitem{habernal-etal-2018-semeval}
Habernal, I., Wachsmuth, H., Gurevych, I., Stein, B.: {S}em{E}val-2018 task 12:
  The argument reasoning comprehension task. In: Proceedings of The 12th
  International Workshop on Semantic Evaluation. pp. 763--772. Association for
  Computational Linguistics, New Orleans, Louisiana (Jun 2018),
  \url{http://www.aclweb.org/anthology/S18-1121}

\bibitem{HW20}
Horn, L.R., Wansing, H.: Negation. In: Zalta, E.N. (ed.) Stanford Encyclopedia
  of Philosophy. Metaphysics Research Lab, Stanford University (2020),
  \url{http://plato.stanford.edu/entries/negation/}

\bibitem{JZ21}
Janssen, T.M.V., Zimmermann, T.E.: Montague semantics. In: Zalta, E.N. (ed.)
  Stanford Encyclopedia of Philosophy. Metaphysics Research Lab, Stanford
  University (2021),
  \url{http://plato.stanford.edu/archives/sum2021/entries/montague-semantics/}

\bibitem{JM+20}
Jim{\'e}nez-Zafra, S.M., Morante, R., Teresa Mart{\'\i}n-Valdivia, M.,
  Ure{\~n}a-L{\'o}pez, L.A.: Corpora annotated with negation: An overview.
  Computational Linguistics  \textbf{46}(1),  1--52 (2020),
  \url{http://doi.org/10.1162/coli_a_00371}

\bibitem{JF20}
Jurafsky, D., H.~James, M.: Vector semantics and embeddings. In: Speech and
  Language Processing: an Introduction to Natural Language Processing,
  Computational Linguistics, and Speech Recognition, chap.~6, pp. 96--126.
  Prentice Hall, Upper Saddle River, N.J, 3rd edn. (2020),
  \url{http://web.stanford.edu/~jurafsky/slp3/ed3book.pdf}, draft

\bibitem{KR93}
Kamp, H., Reyle, U.: From Discourse to Logic: An Introduction to Modeltheoretic
  Semantics of Natural Language, Formal Logic and Discourse Representation
  Theory. Springer, Dordrecht (1993),
  \url{http://www.springer.com/de/book/9780792310280}

\bibitem{KW+20}
Kolhatkar, V., Wu, H., Cavasso, L., Francis, E., Shukla, K., Taboada, M.: The
  {SFU} opinion and comments corpus: A corpus for the analysis of online news
  comments. Corpus Pragmatics  \textbf{4},  155--190 (2020),
  \url{http://link.springer.com/article/10.1007/s41701-019-00065-w}

\bibitem{lapponi2012uio}
Lapponi, E., Velldal, E., {\O}vrelid, L., Read, J.: {UiO~2}: Sequence-labeling
  negation using dependency features. In: Agirre et~al.  \cite{starsem2012},
  pp. 319--327, \url{http://www.aclweb.org/anthology/S12-1042}

\bibitem{lavergne2010practical}
Lavergne, T., Capp{\'e}, O., Yvon, F.: Practical very large scale {CRF}s. In:
  Proceedings of the 48th Annual Meeting of the Association for Computational
  Linguistics. pp. 504--513 (2010),
  \url{http://www.aclweb.org/anthology/P10-1052}

\bibitem{Liu:2004:CMP:1031314.1031373}
Liu, H., Singh, P.: {ConceptNet} -- a practical commonsense reasoning tool-kit.
  BT Technology Journal  \textbf{22}(4),  211--226 (2004),
  \url{http://doi.org/10.1023/B:BTTJ.0000047600.45421.6d}

\bibitem{Lyo77}
Lyons, J.: Semantics, vol.~2. Cambridge University Press, Cambridge, New York,
  Melbourne, Madrid (1977), \url{http://doi.org/10.1017/CBO9780511620614}

\bibitem{copa}
Maslan, N., Roemmele, M., Gordon, A.S.: One hundred challenge problems for
  logical formalizations of commonsense psychology. In: Twelfth International
  Symposium on Logical Formalizations of Commonsense Reasoning, Stanford, CA
  (2015),
  \url{http://www.aaai.org/ocs/index.php/SSS/SSS15/paper/viewFile/10252/10080}

\bibitem{DBLP:journals/cacm/Miller95}
Miller, G.A.: {WordNet}: a lexical database for english. Commun. {ACM}
  \textbf{38}(11),  39--41 (1995), \url{http://doi.org/10.1145/219717.219748}

\bibitem{morante-blanco-2012-sem}
Morante, R., Blanco, E.: *{SEM} 2012 shared task: Resolving the scope and focus
  of negation. In: Agirre et~al.  \cite{starsem2012}, pp. 265--274,
  \url{http://www.aclweb.org/anthology/S12-1035}

\bibitem{morante-daelemans-2012-conandoyle}
Morante, R., Daelemans, W.: {C}onan{D}oyle-neg: Annotation of negation cues and
  their scope in conan doyle stories. In: Proceedings of the Eighth
  International Conference on Language Resources and Evaluation ({LREC}'12).
  pp. 1563--1568. European Language Resources Association (ELRA), Istanbul,
  Turkey (2012),
  \url{http://www.lrec-conf.org/proceedings/lrec2012/pdf/221_Paper.pdf}

\bibitem{storyclozetest}
Mostafazadeh, N., Roth, M., Louis, A., Chambers, N., Allen, J.: {LSDS}em 2017
  shared task: The story cloze test. In: Proceedings of the 2nd Workshop on
  Linking Models of Lexical, Sentential and Discourse-level Semantics. pp.
  46--51 (2017), \url{http://doi.org/10.18653/v1/w17-0906}

\bibitem{Mul14}
Mueller, E.T.: Commonsense Reasoning. Morgan Kaufmann, San Francisco, 2nd edn.
  (2014), \url{http://dl.acm.org/doi/book/10.5555/2821577}

\bibitem{niven-kao-2019-probing}
Niven, T., Kao, H.Y.: Probing neural network comprehension of natural language
  arguments. In: Proceedings of the 57th Annual Meeting of the Association for
  Computational Linguistics. pp. 4658--4664. Association for Computational
  Linguistics, Florence, Italy (Jul 2019),
  \url{http://www.aclweb.org/anthology/P19-1459}

\bibitem{nivre2006maltparser}
Nivre, J., Hall, J., Nilsson, J.: Malt{P}arser: A data-driven parser-generator
  for dependency parsing. In: LREC. vol.~6, pp. 2216--2219 (2006),
  \url{http://aclanthology.org/L06-1084/}

\bibitem{RN+18}
Radford, A., Narasimhan, K., Salimans, T., Sutskever, I.: Improving language
  understanding by generative pre-training. Tech. rep., Open AI (2018),
  \url{http://openai.com/blog/language-unsupervised/}

\bibitem{SSS19a}
Schon, C., Siebert, S., Stolzenburg, F.: The {CoRg} project: Cognitive
  reasoning. KI  \textbf{33}(3),  293--299 (2019),
  \url{http://link.springer.com/article/10.1007/s13218-019-00601-5}

\bibitem{SSS19b}
Siebert, S., Schon, C., Stolzenburg, F.: Commonsense reasoning using theorem
  proving and machine learning. In: Holzinger, A., Kieseberg, P., Tjoa, A.M.,
  Weippl, E. (eds.) Machine Learning and Knowledge Extraction -- Proceedings of
  CD-MAKE 2019. pp. 395--413. LNCS~11713, Springer Nature Switzerland,
  Canterbury, UK (2019), \url{http://doi.org/10.1007/978-3-030-29726-8_25}

\bibitem{SL92}
Simari, G.R., Loui, R.P.: A mathematical treatment of defeasible reasoning and
  its implementation. Artificial Intelligence  \textbf{53}(2-3),  125--157
  (1992), \url{http://doi.org/10.1016/0004-3702(92)90069-A}

\bibitem{speer2017conceptnet}
Speer, R., Chin, J., Havasi, C.: {C}oncept{N}et 5.5: An open multilingual graph
  of general knowledge. In: AAAI Conference on Artificial Intelligence. pp.
  4444--4451 (2017),
  \url{http://aaai.org/ocs/index.php/AAAI/AAAI17/paper/view/14972}

\bibitem{WS16}
Wirth, C.P., Stolzenburg, F.: A series of revisions of {D}avid {P}oole's
  specificity. Annals of Mathematics and Artificial Intelligence
  \textbf{78}(3),  205--258 (2016),
  \url{http://doi.org/10.1007/s10472-015-9471-9}, special issue on {Belief
  Change and Argumentation in Multi-Agent Scenarios}. Issue editors: J{\"u}rgen
  Dix, Sven Ove Hansson, Gabriele Kern-Isberner, Guillermo Simari

\end{thebibliography}

\end{document}